\documentclass[a4,10pt]{article}

\usepackage{multirow}
\usepackage{amsmath}
\usepackage{pgfplots}
\usepackage{algorithm}
\usepackage{algorithmic}
\usepackage{rotating}
\usepackage{graphics}
\usepackage{booktabs}
\usepackage[round]{natbib}
\usepackage{url}

\def\Plus{\texttt{+}}
\def\Minus{\texttt{-}}

\begin{document}
\title{Regularisation of Neural Networks by Enforcing Lipschitz Continuity}
\author{
    Henry Gouk \\
    University of Edinburgh \\
    \texttt{hgouk@inf.ed.ac.uk}
    \and
    Eibe Frank, Bernhard Pfahringer, Michael J. Cree \\
    University of Waikato \\
    \texttt{firstname.lastname@waikato.ac.nz}
}
\date{}

\maketitle

\begin{abstract}
We investigate the effect of explicitly enforcing the Lipschitz continuity of neural networks with respect to their inputs. To this end, we provide a simple technique for computing an upper bound to the Lipschitz constant---for multiple $p$-norms---of a feed forward neural network composed of commonly used layer types. Our technique is then used to formulate training a neural network with a bounded Lipschitz constant as a constrained optimisation problem that can be solved using projected stochastic gradient methods. Our evaluation study shows that the performance of the resulting models exceeds that of models trained with other common regularisers. We also provide evidence that the hyperparameters are intuitive to tune, demonstrate how the choice of norm for computing the Lipschitz constant impacts the resulting model, and show that the performance gains provided by our method are particularly noticeable when only a small amount of training data is available.
\end{abstract}

\section{Introduction}
Supervised learning is primarily concerned with the problem of approximating a function given examples of what output should be produced for a particular input. For the approximation to be of any practical use, it must generalise to unseen data points. Thus, we need to select an appropriate space of functions in which the machine should search for a good approximation, and select an algorithm to search through this space. This is typically done by first picking a large family of models, such as support vector machines or decision trees, and applying a suitable search algorithm. Crucially, when performing the search, regularisation techniques specific to the chosen model family must be employed to combat overfitting. For example, one could limit the depth of decision trees considered by a learning algorithm, or impose probabilistic priors on tunable model parameters.

Regularisation of neural network models is a particularly difficult challenge. The methods that are currently most effective~\citep{srivastava2014, ioffe2015} are heuristically motivated, which can make the process of applying these techniques to new problems nontrivial or unreliable. In contrast, well-understood regularisation approaches adapted from linear models, such as applying an $\ell_2$ penalty term to the model parameters, are known to be less effective than the heuristic approaches~\citep{srivastava2014}. This provides a clear motivation for developing well-founded and effective regularisation methods for neural networks. Following the intuition that functions are considered simpler when they vary at a slower rate, and thus generalise better, we develop a method that allows us to control the Lipschitz constant of a network---a measure of the maximum variation a function can exhibit. Our experiments show that this is a useful inductive bias to impose on neural network models.

One of the prevailing themes in the theoretical work surrounding neural networks is that the magnitude of the weights directly impacts the generalisation gap~\citep{bartlett1998, bartlett2017, neyshabur2017, golowich2018}, with larger weights being associated with poorer relative performance on new data. In several of the most recent works~\citep{bartlett2017, neyshabur2017, golowich2018}, some of the dominant terms in these bounds are equal to the upper bound of the Lipschitz constant of neural networks as we derive it in this paper. While previous works have only considered the Lipschitz continuity of networks with respect to the $\ell_2$ norm, we put a particular emphasis on working with $\ell_1$ and $\ell_\infty$ norms and construct a practical algorithm for constraining the Lipschitz constant of a network during training. The algorithm takes a hyperparameter for each layer that specifies its maximum allowable Lipschitz constant, and these parameters together determine an upper bound on the allowable Lipschitz constant of the entire network. We reuse the same parameter value across multiple layers in our experiments to accelerate the hyperparameter optimisation process.

Several interesting properties of this regularisation technique are demonstrated experimentally. We show that although our algorithm is not competitive when used in isolation, it is highly effective when combined with other commonly used regularisers. Moreover, gains over conventional regularisation approaches are relatively more pronounced when only a small amount of training data is available. We verify that the hyperparameters behave in an intuitive manner: when set to small values, the model capacity is reduced, and as the values of the hyperparameters are increased, the model capacity also increases. Crucially, there is a range of hyperparameter settings where the performance is greater than that of a model trained without our regulariser.

The paper begins with an outline of previous work related to regularisation and the Lipschitz continuity of neural networks in Section~\ref{sec:related}. This is followed by a detailed derivation of the upper bound on the Lipschitz constant of a wide class of feed forward neural networks in Section~\ref{sec:computing}, where we give consideration to multiple choices of vector norms. Section~\ref{sec:regulariser} shows how this upper bound can be used to regularise the neural network in an efficient manner. Experiments showing the utility of this regularisation approach are given in Section~\ref{sec:experiments}, and conclusions are drawn in Section~\ref{sec:conclusion}.

\section{Related work}
\label{sec:related}
One of the most widely applied regularisation techniques currently used for deep networks is dropout~\citep{srivastava2014}. By randomly setting the activations of each hidden unit to zero with some probability, $p$, during training, this method noticeably reduces overfitting for a wide variety of models. Various extensions have been proposed, such as randomly setting weights to zero instead of activations~\citep{wan2013}. Another modification, concrete dropout~\citep{gal2017}, allows one to directly learn the dropout rate, $p$, thus making the search for a good set of hyperparameters easier. \citet{kingma2015} have also shown that the noise level in Gaussian dropout can be learned during optimisation. \citet{srivastava2014} found that constraining the $\ell_2$ norm of the weight vector for each unit in isolation---a technique that they refer to as maxnorm---can improve the performance of networks trained with dropout.

The recent work on optimisation for deep learning has also contributed to our understanding of the generalisation performance of neural networks. Most work in this area aims to be descriptive, rather than prescriptive, in the sense that the focus is on providing explanations for existing heuristic methods as opposed to developing new approaches to improving performance. For example, \citet{hardt2016} quantify the relationship between generalisation error and early stopping. Several papers have shown that the generalisation gap of a neural network is dependent on the magnitude of the weights~\citep{bartlett2017, neyshabur2017, bartlett1998, golowich2018}. Early results, such as \citet{bartlett1998}, present bounds that assume sigmoidal activation functions, but nevertheless relate generalisation to the sum of the absolute values of the weights in the network. More recent work has shown that the product of spectral norms, scaled by various other weight matrix norms, can be used to construct bounds on the generalisation gap. \citet{bartlett2017} scale the spectral norm product by a term related to the element-wise $\ell_1$ norm, whereas \citet{neyshabur2018} use the Frobenius norm. The key quantity used in these bounds is the Lipschitz constant of the parameters of a class of neural networks, which are in turn used in covering number arguments to bound the generalisation performance of models in the hypothesis space.

\citet{neyshabur2018} speculate that Lipschitz continuity with respect to the $\ell_2$ norm alone is insufficient to guarantee generalisation. However, the upper bound presented in Section~\ref{sec:computing} appears in multiple generalisation bounds~\citep{neyshabur2017, bartlett2017, golowich2018}, and we show empirically in this paper that it is an effective aide for controlling the generalisation performance of a deep network. Moreover, the work of \citet{xu2012} demonstrate the concrete link between the Lipschitz constant of a model with respect to its inputs and the resulting generalisation performance. This is accomplished using robustness theory, rather than the tools more typically used in learning theoretic bounds, such as Rademacher complexity and VC dimensions~\citep{shalev2014}. Interestingly, \citet{golowich2018} present a bound on the Rademacher complexity of deep networks that depends only on the product of $\ell_\infty$ operator norms for each weight matrix, which corresponds exactly to the upper bound for the $\ell_\infty$ Lipschitz constant we consider in this paper. This provides yet more evidence that constraining the $\ell_\infty$ Lipschitz constant of a network is a principled method for improving generalisation performance.

\citet{yoshida2017} propose a new regularisation scheme that adds a term to the loss function which penalises the sum of spectral norms of the weight matrices. This is related to but different from what we do in this paper. Firstly, we investigate norms other than $\ell_2$. Secondly, \citet{yoshida2017} use a penalty term, whereas we employ a hard constraint on the induced weight matrix norm, and they penalise the sum of the norms. The Lipschitz constant is determined by the product of operator norms. Finally, they use a heuristic to regularise convolutional layers. Specifically, they compute the largest singular value of a flattened weight tensor, as opposed to deriving the true matrix corresponding to the linear operation performed by convolutional layers, as we do in Section~\ref{sec:convolutional}. Explicitly constructing this matrix and computing its largest singular value---even approximately---would be prohibitively expensive. We provide efficient methods for computing the $\ell_1$ and $\ell_\infty$ norms of convolutional layers exactly, and show how one can approximate the spectral norm efficiently by avoiding the need to explicitly construct the matrix representing the linear operation performed by convolutional layers. \citet{balan2017} provide a means for computing an upper bound to the Lipschitz constant of a restricted class of neural networks known as scattering networks. Although their approach computes tighter bounds than those presented in this paper for the networks they consider, most neural networks that are used in practice do not fit into the scattering network framework.

Enforcing Lipschitz continuity of a network is not only interesting for regularisation. \citet{miyato2018} show that constraining the weights of the discriminator in a generative adversarial network to have a specific spectral norm can improve the quality of generated samples. They use the same technique as~\citet{yoshida2017} to compute these norms, and thus may benefit from the improvements presented in this paper.

Two pieces of related work have been carried out concurrently to this study. \citet{sedghi2018} propose a method for characterising all the singular values of a convolutional layer through the use of Fourier analysis. \citet{tsuzuku2018} propose a similar method for computing the spectral norm of a convolutional layer, with the intention of regularising it in order to improve the adversarial robustness of the resulting model.

\section{Computing the Lipschitz Constant}
\label{sec:computing}
A function, $f: X \to Y$, is said to be Lipschitz continuous if it satisfies

\begin{equation}
\label{eq:lipschitz-def}
D_Y(f(\vec x_1), f(\vec x_2)) \leq k D_X(\vec x_1, \vec x_2) \quad \forall \vec x_1, \vec x_2 \in X,
\end{equation}

\noindent for some real-valued $k \geq 0$, and metrics $D_X$ and $D_Y$. The value of $k$ is known as the Lipschitz constant, and the function can be referred to as being $k$-Lipschitz. Generally, we are interested in the smallest possible Lipschitz constant, but it is not always possible to find it. In this section, we show how to compute an upper bound to the Lipschitz constant of a feed-forward neural network with respect to the input features. Such networks can be expressed as a series of function compositions:

\begin{equation}
\label{eq:feed-forward}
f(\vec x) = (\phi_l \circ \phi_{l-1} \circ ... \circ \phi_1)(\vec x),
\end{equation}

\noindent where each $\phi_i$ is an activation function, linear operation, or pooling operation. A particularly useful property of Lipschitz functions is how they behave when composed: the composition of a $k_1$-Lipschitz function, $f_1$, with a $k_2$-Lipschitz function, $f_2$, is a $k_1k_2$-Lipschitz function. Denoting the Lipschitz constant of some function, $f$, as $L(f)$, repeated application of this composition property yields the following upper bound on the Lipschitz constant for the entire feed-forward network:

\begin{equation}
\label{eq:lipschitz-product}
L(f) \leq \prod_{i = 1}^{l} L(\phi_i).
\end{equation}

Thus, we can compute the Lipschitz constants of each layer in isolation and combine them in a modular way to establish an upper bound on the constant of the entire network. It is important to note that $k_1k_2$ will not necessarily be the smallest Lipschitz constant of $(f_2 \circ f_1)$, even if $k_1$ and $k_2$ are individually the best Lipschitz constants of $f_1$ and $f_2$, respectively. It is possible in theory that a tighter upper bound can be obtained by considering the entire network as a whole rather than each layer in isolation. In the remainder of this section, we derive closed form expressions for the Lipschitz constants of common layer types when $D_X$ and $D_Y$ correspond to $\ell_1$, $\ell_2$, or $\ell_\infty$ norms respectively. As we will see in Section~\ref{sec:regulariser}, Lipschitz constants with respect to these norms can be constrained efficiently.

\subsection{Fully Connected Layers}
\label{sec:fully-connected}
A fully connected layer, $\phi^{fc}(\vec x)$, implements an affine transformation parameterised by a weight matrix, $W$, and a bias vector, $\vec b$:

\begin{equation}
\label{eq:fully-connected}
\phi^{fc}(\vec x) = W \vec x + \vec b.
\end{equation}

Others have already established that, under the $\ell_2$ norm, the Lipschitz constant of a fully connected layer is given by the spectral norm of the weight matrix~\citep{miyato2018, neyshabur2017}. We provide a slightly more general formulation that will prove to be more useful when considering other $p$-norms. We begin by plugging the definition of a fully connected layer into the definition of Lipschitz continuity:

\begin{equation}
\|(W \vec x_1 + \vec b) - (W \vec x_2 + \vec b)\|_p \leq k \|\vec x_1 - \vec x_2\|_p.
\end{equation}

\noindent By setting $\vec a = \vec x_1 - \vec x_2$ and simplifying the expression slightly, we arrive at

\begin{equation}
\|W \vec a\|_p \leq k \|\vec a\|_p,
\end{equation}

\noindent which, assuming $\vec x_1 \neq \vec x_2$, can be rearranged to

\begin{equation}
\frac{\|W \vec a\|_p}{\|\vec a\|_p} \leq k, \quad \vec a \neq 0.
\end{equation}

The smallest Lipschitz constant is therefore equal to the supremum of the left-hand side of the inequality,

\begin{equation}
L(\phi^{fc}) = \sup_{\vec a \neq 0} \frac{\|W \vec a\|_p}{\|\vec a\|_p},
\end{equation}

\noindent which is the definition of the operator norm of $W$.

For the $p$-norms we consider in this paper, there exist efficient algorithms for computing operator norms on relatively large matrices. Specifically, for $p=1$, the operator norm is the maximum absolute column sum norm; for $p=\infty$, the operator norm is the maximum absolute row sum norm. The time required to compute both of these norms is linearly related to the number of elements in the weight matrix. When $p = 2$, the operator norm is given by the largest singular value of the weight matrix---the spectral norm---which can be approximated relatively quickly using a small number of iterations of the power method.

\subsection{Convolutional Layers}
\label{sec:convolutional}
Convolutional layers, $\phi^{conv}(X)$, also perform an affine transformation, but it is usually more convenient to express the computation in terms of discrete convolutions and point-wise additions. For a convolutional layer, the $i$-th output feature map is given by

\begin{equation}
\label{eq:conv-layer}
\phi^{conv}_i(X) = \sum_{j=1}^{M_{l-1}} F_{i,j} \ast X_j + B_i,
\end{equation}

\noindent where each $F_{i,j}$ is a filter, each $X_j$ is an input feature map, $B_i$ is an appropriately shaped bias tensor exhibiting the same value in every element, and the previous layer produced $M_{l-1}$ feature maps.

The convolutions in Equation~\ref{eq:conv-layer} are linear operations, so one can exploit the isomorphism between linear operations and square matrices of the appropriate size to reuse the matrix norms derived in Section~\ref{sec:fully-connected}. To represent a single convolution operation as a matrix--vector multiplication, the input feature map is serialised into a vector, and the filter coefficients are used to construct a doubly block circulant matrix. Due to the structure of doubly block circulant matrices, each filter coefficient appears in each column and row of this matrix exactly once. Consequently, the $\ell_1$ and $\ell_\infty$ operator norms are the same and given by $\|F_{i,j}\|_1$, the sum of the absolute values of the filter coefficients used to construct the matrix.

Summing over several different convolutions associated with different input feature maps and the same output feature map, as done in Equation~\ref{eq:conv-layer}, can be accomplished by horizontally concatenating matrices. For example, suppose $V_{i,j}$ is a matrix that performs a convolution of $F_{i,j}$ with the $j$-th feature map serialised into a vector. Equation~\ref{eq:conv-layer} can now be rewritten in matrix form as

\begin{equation}
\phi^{conv}_i(\vec x) = \lbrack V_{1,1} \  V_{1,2} \  ... \  V_{1,M_{l-1}} \rbrack \vec x + \vec b_i,
\end{equation}

\noindent where the inputs and biases, previously represented by $X$ and $B_i$, have been serialised into vectors $\vec x$ and $\vec b_i$, respectively. The complete linear transformation, $W$, performed by a convolutional layer to generate $M_l$ output feature maps can be constructed by adding additional rows to the block matrix:

\begin{equation}
\label{eq:true-conv-op}
W =
\begin{bmatrix}
V_{1,1} & \hdots & V_{1,M_{l-1}} \\
\vdots & \ddots & \\
V_{M_l,1} &  & V_{M_l,M_{l-1}}
\end{bmatrix}.
\end{equation}

To compute the $\ell_1$ and $\ell_\infty$ operator norms of $W$, recall that the operator norm of $V_{i,j}$ for $p \in \{1, \infty\}$ is $\|F_{i,j}\|_1$. A second matrix, $W^\prime$, can be constructed from $W$, where each block, $V_{i,j}$, is replaced with the corresponding operator norm, $\|F_{i,j}\|_1$. Each of these operator norms can be thought of as a partial row or column sum for the original matrix, $W$. Now, based on the discussion in Section~\ref{sec:fully-connected}, the $\ell_1$ operator norm is given by

\begin{equation}
\|W\|_1 = \max_{j} \sum_{i=1}^{M_l} \|F_{i,j}\|_1,
\end{equation}

\noindent and the $\ell_\infty$ operator norm is given by

\begin{equation}
\|W\|_\infty = \max_{i} \sum_{j=1}^{M_{l-1}} \|F_{i,j}\|_1,
\end{equation}

We now consider the spectral norm for convolutional layers. \citet{yoshida2017} and \citet{miyato2018} both investigate the effect of penalising or constraining the spectral norm of convolutional layers by reinterpreting the weight tensor of a convolutional layer as a matrix,

\begin{equation}
U = 
\begin{bmatrix}
\vec u_{1,1} & \hdots & \vec u_{1,M_{l-1}} \\
\vdots & \ddots & \\
\vec u_{M_l,1} &  & \vec u_{M_l,M_{l-1}}
\end{bmatrix},
\end{equation}

\noindent where each $\vec u_{i,j}$ contains the elements of the corresponding $F_{i,j}$ serialised into a row vector. They then proceed to compute the spectral norm of $U$, rather than computing the spectral norm of $W$, given in Equation~\ref{eq:true-conv-op}. As \citet{cisse2017} and \citet{tsuzuku2018} show, this only computes a loose upper bound of the true spectral norm.

Explicitly constructing $W$ and applying a conventional singular value decomposition to compute the spectral norm of a convolutional layer is infeasible, but we show how the power method can be adapted to use standard convolutional network primitives to compute it efficiently. Consider the usual process for computing the largest singular value of a square matrix using the power method, provided in Algorithm~\ref{alg:power-method}. The expression of most interest to us is inside the for loop, namely

\begin{equation}
\vec x_{i} = W^{T}W\vec x_{i-1},
\end{equation}

\noindent which, due to the associativity of matrix multiplication, can be broken down into two steps:

\begin{equation}
\label{eq:power-forward}
\vec x_i^{\prime} = W\vec x_{i-1}
\end{equation}

\noindent and

\begin{equation}
\label{eq:power-backward}
\vec x_{i} = W^{T}\vec x_{i}^{\prime}.
\end{equation}

When $W$ is the matrix in Equation~\ref{eq:true-conv-op}, the expressions given in Equations~\ref{eq:power-forward} and \ref{eq:power-backward} correspond to a forward propagation and a backwards propagation through a convolutional layer, respectively. Thus, if we replace these matrix multiplication with convolution and transposed convolution operations respectively, as implemented in many deep learning frameworks, the spectral norm can be computed efficiently. Note that only a single vector must undergo the forward and backward propagation operations, rather than an entire batch of instances. This means, for most cases, only a small increase in runtime will be incurred by using this method. It also automatically takes into account the padding and stride hyperparameters used by the convolutional layer. In contrast to the reshaping method used by \citet{yoshida2017} and \citet{miyato2018}, the approach we use is capable of computing the spectral norm of a convolutional layer exactly if it is run until convergence.

\begin{algorithm}[t]
\caption{Power method for producing the largest singular value, $\sigma_{max}$, of a non-square matrix, $W$.}
\label{alg:power-method}
\begin{algorithmic}
\STATE Randomly initialise $\vec x_0$
\FOR{$i = 1$ \textbf{to} $n$}
	\STATE $\vec x_i \gets W^{T}W\vec x_{i-1}$
\ENDFOR
\STATE $\sigma_{max} \gets \frac{\|W\vec x_n\|_2}{\|\vec x_n\|_2}$
\end{algorithmic}
\end{algorithm}

\subsection{Pooling Layers and Activation Functions}
Computing Lipschitz constants for pooling layers and activations is trivial for commonly used pooling operations and activation functions. Most common activation functions and pooling operations are, at worst, 1-Lipschitz with respect to all $p$-norms. For example, the maximum absolute sub-gradient of the ReLU activation function is 1, which means that ReLU operations have a Lipschitz constant of one. A similar argument yields that the Lipschitz constant of max pooling layers is one. The Lipschitz constant of the softmax is one~\citep{gao2017}.

\subsection{Residual Connections}
Recently developed feed-forward architectures often include residual connections between non-adjacent layers~\citep{he2016}. These are most commonly used to construct structures known as residual blocks:

\begin{equation}
\label{eq:residual-block}
\phi^{res}(\vec x) = \vec x + (\phi_{j+n} \circ ... \circ \phi_{j+1})(\vec x),
\end{equation}

\noindent where the function composition may contain a number of different linear transformations and activation functions. In most cases, the composition is formed by two convolutional layers, each preceded by a batch normalisation layer\footnote{We discuss batch normalisation and the corresponding Lipschitz constant in Section~\ref{sec:batchnorm} below.} and a ReLU function. While networks that use residual blocks still qualify as feed-forward networks, they no longer conform to the linear chain of function compositions we formalised in Equation~\ref{eq:feed-forward}. Fortunately, networks with residual connections are usually built by composing a linear chain of residual blocks of the form given in Equation~\ref{eq:residual-block}. Hence, the Lipschitz constant of a residual network will be the product of Lipschitz constants for each residual block. Each block is a sum of two functions (see Equation 18). Thus, for a $k_1$-Lipschitz function, $f_1$, and a $k_2$-Lipschitz function, $f_2$, we are interested in the Lipschitz constant of their sum:

\begin{equation}
\|(f_1(\vec x_1) + f_2(\vec x_1)) - (f_1(\vec x_2) + f_2(\vec x_2))\|_p
\end{equation}

\noindent which can be rearranged to

\begin{equation}
\label{eq:associative-rearrange}
\|(f_1(\vec x_1) - f_1(\vec x_2)) + (f_2(\vec x_1) - f_2(\vec x_2))\|_p.
\end{equation}

The subadditivity property of norms and the Lipschitz constants of $f_1$ and $f_2$ can then be used to bound Equation~\ref{eq:associative-rearrange} from above:

\begin{align}
\|(f_1(\vec x_1) - f_1(\vec x_2)) + (f_2(\vec x_1) - f_2(\vec x_2))\|_p &\leq \|f_1(\vec x_1) - f_1(\vec x_2)\|_p + \|f_2(\vec x_1) - f_2(\vec x_2)\|_p\\
&\leq k_1 \|\vec x_1 - \vec x_2\|_p + k_2 \|\vec x_1 - \vec x_2\|_p\\
&= (k_1 + k_2) \|\vec x_1 - \vec x_2\|_p.
\end{align}

Thus, we can see that the Lipschitz constant of the addition of two functions is bounded from above by the sum of their Lipschitz constants. Setting $f_1$ to be the identity function and $f_2$ to be a linear chain of function compositions, we arrive at the definition of a residual block as given in Equation~\ref{eq:residual-block}. Noting that the Lipschitz constant of the identity function is one, we can see that the Lipschitz constant of a residual block is bounded by

\begin{equation}
L(\phi^{res}) \leq 1 + \prod_{i = j+1}^{j+n} L(\phi_i),
\end{equation}

\noindent where the property given in Equation~\ref{eq:lipschitz-product} has been applied to the function compositions.

\section{Constraining the Lipschitz Constant}
\label{sec:regulariser}
The assumption motivating our work is that adjusting the Lipschitz constant of a feed-forward neural network controls how well the model will generalise to new data. Using the composition property of Lipschitz functions, we have shown that the Lipschitz constant of a network is the product of the Lipschitz constants associated with its layers. Thus, controlling the Lipschitz constant of a network can be accomplished by constraining the Lipschitz constant of each layer in isolation. This can be achieved by performing constrained optimisation when training the network. In practice, we pick a single hyperparameter, $\lambda$, and use it to control the upper bound of the Lipschitz constant for each layer. This means the network as a whole will have a Lipschitz constant less than or equal to $\lambda^d$, where $d$ is the depth of the network.

The easiest way to adapt existing deep learning methods to allow for constrained optimisation is to introduce a projection step and perform a variant of the projected stochastic gradient method. In our particular problem, because each parameter matrix is constrained in isolation, it is straightforward to project any infeasible parameter values back into the set of feasible matrices. Specifically, after each weight update step, we must check that none of the weight matrices (including the filter banks in the convolutional layers) are violating the constraint on the Lipschitz constant. If the weight update has caused a weight matrix to leave the feasible set, we must replace the resulting matrix with the closest matrix that does lie in the feasible set. This can all be accomplished with the projection function

\begin{equation}
\label{eq:projection}
\pi(W, \lambda) = \frac{1}{\max(1, \frac{\|W\|_p}{\lambda})} W,
\end{equation}

\noindent which will leave the matrix untouched if it does not violate the constraint, and project it back to the closest matrix in the feasible set if it does. Closeness is measured by the matrix distance metric induced by taking the operator norm of the difference between two matrices. This will work with any valid operator norm because all norms are absolutely homogeneous~\cite{pugh2002}. In particular, it will work with the operator norms with $p \in \{1, 2, \infty\}$, which can be computed using the approaches outlined in Section~\ref{sec:computing}.

Pseudocode for this projected gradient method is given in Algorithm~\ref{alg:projected-grad}. We have observed fast convergence when using the Adam update rule~\citep{kingma2014}, but other variants of the stochastic gradient method also work. For example, in our experiments, we show that stochastic gradient descent with Nesterov's momentum is compatible with our approach.

\begin{algorithm}[t]
\caption{Projected stochastic gradient method to 
optimise a neural network subject to the Lipschitz Constant Constraint (LCC). $W_{1:l}$ is used to refer to all $W_i$ for $i \in \{1, ..., l\}$.}
\label{alg:projected-grad}
\begin{algorithmic}
\STATE $t \gets 0$
\WHILE{$W_{1:l}^{(t)}$ not converged}
	\STATE $t \gets t + 1$
	\STATE $g_{1:l}^{(t)} \gets \nabla_{W_{1:l}} f(W_{1:l}^{(t-1)})$
	\STATE $\widehat{W}_{1:l}^{(t)} \gets update(W_{1:l}^{(t-1)}, g_{1:l}^{(t)})$
	\FOR{$i = 1$ \textbf{to} $l$}
		\STATE $W_{i}^{(t)} \gets \pi(\widehat{W}_{i}^{(t)}, \lambda)$
	\ENDFOR
\ENDWHILE
\end{algorithmic}
\end{algorithm}

\subsection{Stability of $p$-norm Estimation}
A natural question to ask is which $p$-norm should be chosen when using the training procedure given in Algorithm~\ref{alg:projected-grad}. The Euclidean (i.e., spectral) norm is often seen as the default choice, due to its special status when talking about distances in the real world. Like~\citet{yoshida2017}, we use the power method to estimate the spectral norms of the linear operations in deep networks. The convergence rate of the power method is related to the ratio of the two largest singular values, $\frac{\sigma_2}{\sigma_1}$~\citep{larson2016}. If the two largest singular values are almost the same, it will converge very slowly. Because each iteration of the power method for computing the spectral norm of a convolutional layer requires both forward propagation and backward propagation, it is only feasible to perform a small number of iterations before one will notice an impact in the training speed. However, regardless of the quality of the approximation, we can be certain that it does not overestimate the true norm: the expression in the final line of Algorithm~\ref{alg:power-method} is maximised when $\vec x_n$ is the first eigenvector of $W$. Therefore, if the algorithm has not converged, $\vec x_n$ will not be a singular vector of $W$ and our approximation of $\sigma_{max}$ will be an underestimate.

In contrast to the spectral norm, we compute the values of the $\ell_1$ and $\ell_\infty$ norms exactly, in time that is linear in the number of weights in a layer, so it always comprises a relatively small fraction of the overall runtime for training the network. Of course, it may be the case that the $\ell_1$ and $\ell_\infty$ constraints do not provide as suitable an inductive bias as the $\ell_2$ constraint. This is something we investigate in our experimental evaluation.

\subsection{Compatibility with Batch Normalisation}
\label{sec:batchnorm}
Constraining the Lipschitz constant of the network will have an impact on the magnitude of the activations produced by each layer, which is what batch normalisation attempts to explicitly control~\citep{ioffe2015}. Thus, we consider whether batch normalisation is compatible with our Lipschitz Constant Constraint (LCC) regulariser. Batch normalisation can be expressed as

\begin{equation}
\phi^{bn}(\vec x) = \text{diag}\Bigg(\frac{\vec \gamma}{\sqrt{\text{Var}\lbrack \vec x \rbrack}}\Bigg) (\vec x - \text{E}\lbrack \vec x \rbrack) + \vec \beta,
\end{equation}

\noindent where $\text{diag}(\cdot)$ denotes a diagonal matrix, and $\vec \gamma$ and $\vec \beta$ are learned parameters. This can be seen as performing an affine transformation with a linear transformation term

\begin{equation}
\text{diag}\Bigg(\frac{\vec \gamma}{\sqrt{\text{Var}\lbrack \vec x \rbrack}}\Bigg) \vec x.
\end{equation}

Based on the operator norm of this diagonal matrix, the Lipschitz constant of a batch normalisation layer, with respect to the three $p$-norms we consider, is given by

\begin{equation}
\label{eq:bn-lipschitz}
L(\phi^{bn}) = \max_i \Bigg|\frac{\vec \gamma_i}{\sqrt{\text{Var}\lbrack \vec x_i \rbrack}}\Bigg|.
\end{equation}

Thus, when using batch normalisation in conjunction with our technique, the $\vec \gamma$ parameter must also be constrained. This is accomplished by using the expression in Equation~\ref{eq:bn-lipschitz} to compute the operator norm in the projection function given in Equation~\ref{eq:projection}. In practice, when training the network with minibatch gradient descent, we use a moving average estimate of the variance for performing the projection, rather than the variance computed solely on the current minibatch of training examples. This is done because the minibatch estimates of the mean and variance can be quite noisy.

\subsection{Interaction with Dropout}
\label{sec:dropout}
In the standard formulation of dropout, one corrupts activations during training by performing pointwise multiplication with vectors of Bernoulli random variables. As a consequence, when making a prediction at test time---when units are not dropped out---the activations must be scaled by the probability that they remained uncorrupted during training. This means the activation magnitude at both test time and training time is approximately the same. The majority of modern neural networks make extensive use of rectified linear units, or similar activation functions that are also homogeneous. This implies that scaling the activations at test time is equivalent to scaling the weight matrices in the affine transformation layers. By definition, this will also scale the operator norm, and therefore the Lipschitz constant, of that layer. As a result, one may expect that when using our technique in conjunction with dropout, the $\lambda$ hyperparameter will need to be increased in order to maintain the desired Lipschitz constant. Note that this does not imply that the optimal value for $\lambda$, from the point of view of generalisation performance, can be found by performing hyperparameter optimisation without dropout, and then dividing the best $\lambda$ found on the validation set by one minus the desired dropout rate: the change in optimisation dynamics and regularisation properties of dropout make it difficult to predict analytically how these two methods interact when considering generalisation performance.

\section{Experiments}
\label{sec:experiments}
The experiments in this section aim to answer several questions about the behaviour of the Lipschitz Constant Constraint (LCC) regularisation scheme presented in this paper. The question of most interest is how well this regularisation technique compares to related regularisation methods, in terms of accuracy measured on held-out data. In addition to this, experiments are performed that demonstrate how sensitive the method is to the choice of values of the $\lambda$ hyperparameters, how it interacts with existing regularisation methods, and how the additional inductive bias imposed on the learning system impacts the sample efficiency.

Several different network architectures are employed in the experiments. Specifically, fully connected multi-layer perceptrons, VGG-style convolutional networks, and networks with residual connections are used. This is to ensure that the regularisation method works for a broad range of feed-forward architectures. SGD with Nesterov momentum is used for training networks with residual connections, and the Adam optimiser~\citep{kingma2014} is used otherwise. Batch normalisation is used in all networks to accelerate training. All regularisation hyperparameters for the convolutional networks were optimised on a per-layer type basis using the hyperopt package\footnote{\url{https://github.com/hyperopt/hyperopt}} of~\citet{bergstra2015}. Separate dropout, spectral decay, and $\lambda$ hyperparameters were optimised for fully connected and convolutional layers. All network weights were initialised using the method of \citet{glorot2010}, and the estimated accuracy reported in all tables is the mean of five networks that were each initialised using different seeds, unless stated otherwise. The standard deviation is also reported to give an idea of how robust different regularisers are to different initialisations. The code for running these experiments is available online.\footnote{\url{https://github.com/henrygouk/keras-lipschitz-networks}}

\subsection{CIFAR-10}

The CIFAR-10 dataset~\citep{krizhevsky2009} contains 60,000 tiny images, each belonging to one of 10 classes. The experiments in this section follow the common protocol of using 10,000 of the images in the 50,000 image training set for tuning the model hyperparameters. Two network architectures are considered for this dataset: a VGG19-style network~\citep{simonyan2014}, resized to be compatible with the $32 \times 32$ pixel images in CIFAR-10, and a Wide Residual Network (WRN)~\citep{zagoruyko2016}. All experiments on this dataset utilise data augmentation in the form of random crops and horizontal flips, and the image intensities were rescaled to fall into the $\lbrack -1, 1 \rbrack$ range. Each VGG network is trained for 140 epochs using the Adam optimiser~\citep{kingma2014}. The initial learning rate is set to $10^{-4}$ and decreased by a factor of 10 after epoch 100 and epoch 120. The WRNs are trained for a total of 200 epochs using the stochastic gradient method with Nesterov's momentum. The learning rate was initialised to $0.1$, and decreased by a factor of 5 at epochs 60, 120, and 160.

The performance of LCC is compared to dropout and the spectral decay method of~\cite{yoshida2017}. Dropout is a widely used regularisation method, often acting as key components of state-of-the-art models~\citep{simonyan2014,he2016,zagoruyko2016}, and the spectral decay method has a similar goal to the $\ell_2$ instantiation of our method: encouraging the spectral norm of the weight matrices to be small. For this particular experiment, each regulariser is considered in isolation, but we also consider combinations of LCC and spectral decay with dropout. Results are given in Table~\ref{tbl:cifar10-exps}. Interestingly, the performance of the VGG network varies considerably more than that of the Wide Residual Network. VGG networks see the most benefit from LCC-$\ell_2$, but dropout and spectral decay do not provide any noticeable inmprovement in performance. Combining dropout with the other methods is not an effective strategy on this dataset. In the case of WRNs, LCC performs similarly to dropout and marginally better than spectral decay, but there is little separation between methods on this dataset.

\begin{table}[t]
\center
\caption{Performance of VGG19 and WRN-16-10 networks trained with spectral decay, dropout, LCC, and combinations thereof on CIFAR-10. LCC-$\ell_p$ denotes the Lipschitz Constant Constraint method for a given $p$-norm.}
\label{tbl:cifar10-exps}
\begin{tabular}{lcc}
\toprule
    Method & VGG19 & WRN-16-10 \\
\midrule
    None                        & 90.43$\pm$0.17 & 95.13$\pm$0.17 \\
    Dropout                     & 90.46$\pm$0.10 & 95.46$\pm$0.20 \\
    Spectral Decay				& 90.14$\pm$0.16 & 95.21$\pm$0.08 \\
    LCC-$\ell_1$                & 92.48$\pm$0.13 & 95.34$\pm$0.21 \\
    LCC-$\ell_2$                & 92.57$\pm$0.28 & 95.32$\pm$0.15 \\
    LCC-$\ell_\infty$           & 91.64$\pm$0.20 & \textbf{95.82$\pm$0.19} \\
    Dropout + Spectral Decay    & 90.29$\pm$0.17 & 95.22$\pm$0.08 \\
    Dropout + LCC-$\ell_1$      & 91.72$\pm$0.17 & 95.47$\pm$0.15 \\
    Dropout + LCC-$\ell_2$      & 91.23$\pm$0.24 & 95.57$\pm$0.15 \\
    Dropout + LCC-$\ell_\infty$ & \textbf{92.71$\pm$0.29} & 95.68$\pm$0.11 \\
\bottomrule
\end{tabular}
\end{table}

\subsection{CIFAR-100}
\label{sec:cifar100-exps}
CIFAR-100, like CIFAR-10, is a dataset of 60,000 tiny images, but contains 100 classes rather than 10. The same data augmentation methods used for CIFAR-10 are also used for training models on CIFAR-100---random crops and horizontal flips. Once again, WRNs and VGG19-style networks ared trained on this dataset. The learning rate schedules used in the CIFAR-10 experiments also worked well on this dataset, which is not surprising given their similarities. However, the regularisation hyperparameters were optimised specifically for CIFAR-100. The results for the VGG and WRN models are given in Table~\ref{tbl:cifar100-exps}.

\begin{table}[t]
\center
\caption{Performance of networks trained with spectral decay, dropout, LCC, and combinations thereof on CIFAR-100. LCC-$\ell_p$ denotes our Lipschitz Constant Constraint method for some given $p$-norm.}
\label{tbl:cifar100-exps}
\begin{tabular}{lcc}
\toprule
Method & VGG19 & WRN-16-10 \\
\midrule
	None 						& 65.46$\pm$0.43 & 77.94$\pm$0.33 \\
	Dropout 					& 66.75$\pm$0.40 & 77.98$\pm$0.24 \\
	Spectral Decay				& 65.32$\pm$0.24 & 77.93$\pm$0.20 \\
	LCC-$\ell_1$ 				& 69.59$\pm$0.29 & 78.16$\pm$0.04 \\
	LCC-$\ell_2$ 				& 68.25$\pm$0.38 & 79.00$\pm$0.33 \\
	LCC-$\ell_\infty$ 			& 69.16$\pm$0.22 & 79.39$\pm$0.28 \\
	Dropout + Spectral Decay    & 66.97$\pm$0.24 & 77.70$\pm$0.33 \\
	Dropout + LCC-$\ell_1$ 		& 70.17$\pm$0.21 & 79.08$\pm$0.11 \\
	Dropout + LCC-$\ell_2$ 		& \textbf{71.76$\pm$0.26} & \textbf{79.45$\pm$0.26} \\
	Dropout + LCC-$\ell_\infty$ & 69.25$\pm$0.43 & 78.17$\pm$1.86 \\
\bottomrule
\end{tabular}
\end{table}

It can be seen that the Lipschitz-based regularisation scheme is an effective technique for improving generalisation of networks both with and without residual connections. The results on CIFAR-100 follow a similar trend to those observed on CIFAR-10: LCC performs the best, dropout provides a small increase in performance over no regularisation, and combining dropout other approaches can sometimes provide a small boost in accuracy. Spectral decay performs noticeably worse than LCC-$\ell_2$, often having comparable performance to no regularisation.

\subsection{MNIST and Fashion-MNIST}
The Fashion-MNIST dataset~\citep{xiao2017} is designed as a more challenging drop-in replacement for the original MNIST dataset of hand-written digits~\citep{lecun1998}. Both contain $70,000$ greyscale images labelled with one of 10 possible classes. The last 10,000 instances are used as the test set. The final 10,000 instances in the training set are used for measuring performance when optimising the regularisation hyperparameters. In these experiments, small convolutional networks are trained on both of these datasets with different combinations of regularisers. The networks contain only two convolutional layers, each consisting of $5 \times 5$ kernels, and both layers are followed by $2 \times 2$ max pooling layers. The first layer has 64 feature maps, and the second has 128. These layers feed into a fully connected layer with 128 units, which is followed by the output layer with 10 units. ReLU activations are used for all hidden layers, and each model is trained for 60 epochs using Adam~\citep{kingma2014}. The learning rate was started at $10^{-4}$ and decreased by a factor of 10 at the fiftieth epoch.

The test accuracies for each of the models trained on these datasets are given in Table~\ref{tbl:mnist}. For both datasets, dropout and spectral decay decrease performace, whereas LCC-$\ell_1$ results in a consistent performance increase.

\begin{table}[t]
\centering
\caption{Test accuracies of the small convolutional networks trained with spectral decay, dropout, LCC, and combinations thereof on the MNIST and Fashion-MNSIT datasets.}
\label{tbl:mnist}
\begin{tabular}{lcc}
\toprule
Method & MNIST & Fashion-MNIST \\
\midrule
None                        & 99.29$\pm$0.03 & 92.54$\pm$0.10 \\
Dropout                     & 98.93$\pm$0.17 & 91.68$\pm$0.17 \\
Spectral Decay              & 99.28$\pm$0.07 & 92.59$\pm$0.03 \\
LCC-$\ell_1$                & 99.41$\pm$0.05 & 93.06$\pm$0.15 \\
LCC-$\ell_2$                & 99.41$\pm$0.05 & 92.62$\pm$0.18 \\
LCC-$\ell_\infty$           & 99.32$\pm$0.09 & 92.87$\pm$0.12 \\
Dropout + Spectral Decay    & 98.85$\pm$0.15 & 91.86$\pm$0.18 \\
Dropout + LCC-$\ell_1$      & 99.35$\pm$0.08 & \textbf{93.23$\pm$0.23} \\
Dropout + LCC-$\ell_2$      & \textbf{99.42$\pm$0.10} & 91.71$\pm$0.38 \\
Dropout + LCC-$\ell_\infty$ & 99.36$\pm$0.04 & 92.75$\pm$0.25 \\
\bottomrule
\end{tabular}
\end{table}

\subsection{Street View House Numbers}
The Street View House Numbers dataset contains over $600,000$ images of digits extracted from Google's Street View platform. Each image contains three colour channels and has a resolution of $32 \times 32$ pixels. As with the previous datasets, the only preprocessing performed is to rescale the input features to the range $\lbrack -1, 1\rbrack$. However, in contrast to the experiments on CIFAR-10 and CIFAR-100, no data augmentation is performed while training on this dataset. The first network architecture used for this dataset, which follows a VGG-style structure, is comprised of four conv--conv--maxpool blocks with 64, 128, 192, and 256 feature maps, respectively. This is followed by two fully connected layers, each with 512 units, and then the logistic regression layer. Due to the large training set size, it is only necessary to train for 20 epochs. The Adam optimiser~\citep{kingma2014} is used with an initial learning rate of $10^{-4}$, which is decreased by a factor of 10 at epochs 15 and 18. Small WRN models are also trained on this dataset. Once again, due to the large size of the training set, it is sufficient to only train each network for 20 epochs in total. Therefore, compared to the WRNs trained on CIFAR-10 and CIFAR-100, a compressed learning rate schedule is used. The learning rate is started at 0.1, and is decreased by a factor of 5 at epochs 6, 12, and 16. Measurements of the test set performance for each of the models trained on SVHN are provided in Table~\ref{tbl:svhn}.

For VGG, using dropout in conjunction with other approaches results in the best performance, but in isolation is not effective. LCC improves accuracy in both the VGG and WRN models, whereas spectral decay does not help for either.

\begin{table}[t]
\center
\caption{Prediction accuracy of VGG-style and WRN-16-4 networks trained with spectral decay, dropout, LCC, and combinations thereof on the SVHN dataset.}
\label{tbl:svhn}
\begin{tabular}{lcc}
\toprule
Method & VGG & WRN-16-4 \\
\midrule
None 						& 96.90$\pm$0.05 & 97.97$\pm$0.04 \\
Dropout 					& 96.98$\pm$0.10 & 98.23$\pm$0.05 \\
Spectral Decay				& 96.88$\pm$0.04 & 98.02$\pm$0.04 \\
LCC-$\ell_1$ 				& 97.17$\pm$0.09 & 98.00$\pm$0.06 \\
LCC-$\ell_2$ 				& 96.94$\pm$0.04 & 97.93$\pm$0.07 \\
LCC-$\ell_\infty$ 			& 97.35$\pm$0.03 & 98.03$\pm$0.05 \\
Dropout + Spectral Decay    & 97.10$\pm$0.06 & 98.15$\pm$0.04 \\
Dropout + LCC-$\ell_1$ 		& 97.30$\pm$0.07 & 98.21$\pm$0.02 \\
Dropout + LCC-$\ell_2$ 		& \textbf{97.73$\pm$0.30} & 98.17$\pm$0.05 \\
Dropout + LCC-$\ell_\infty$ & 97.32$\pm$0.06 & \textbf{98.24$\pm$0.06} \\
\bottomrule
\end{tabular}
\end{table}

\subsection{Scaled ImageNet Subset (SINS-10)}
The SINS-10 dataset is a collection of 100,000 images taken from ImageNet by~\citet{gouk2019}. Each image in this dataset is $96 \times 96$ pixels and is labelled with one of 10 classes. What makes this dataset distinct from other commonly used image classification benchmarks is that it is divided into 10 non-overlapping and equal sized predefined folds. Within each fold, 9,000 images are used for training and 1,000 are used for testing. By gathering multiple estimates of algorithm performance, one can perform hypothesis tests to determine statistically significant differences between methods.

The experiments conducted on SINS-10 in this paper make use of the same VGG-style and WRN network architectures used for the SVHN experiments. However, because each fold of the SINS-10 dataset has many fewer instances than SVHN, the number of epochs and learning rate schedules are changed. The VGG networks are trained for a total of 60 epochs, beginning with a learning rate of $10^{-4}$ that is decreased by a factor of 10 at epochs 40 and 50. The WRN models are trained for 100 epochs each, with a starting learning rate of $0.1$ that is decreased by a factor of five at epochs 30, 60 and 80. The regularisation hyperparameters are optimised on a per-fold basis. The final 1,000 instances of the training set are repurposed as a validation set to determine the quality of a given hyperparameter setting. The results for these experiments are given in Table~\ref{tbl:sins10-vgg} for the VGG models, and Table~\ref{tbl:sins10-wrn} for the wide residual network models. Hypothesis tests are carried out using a paired $t$-test to determine whether using the regulariser improves performance. In the case where a method is used in conjunction with dropout, the hypothesis test compares the performance of the combination with that of dropout along.

\begin{table}[t]
    \center
    \caption{Prediction accuracies of VGG-style networks trained with spectral decay, dropout, batchnorm, LCC, and combinations thereof on the SINS-10 dataset. The \Plus/\Minus \, column indicates whether adding LCC to the combination of regularisers results in a statistically significant improvement  or degradation in performance at the 95\% confidence level.}
    \label{tbl:sins10-vgg}
    \begin{tabular}{lcc}
        \toprule
        Method & VGG & \Plus/\Minus \\
        \midrule
        None                        & 63.73$\pm$1.18 & \\
        Dropout                     & 68.65$\pm$1.00 & \Plus \\
        Spectral Decay              & 63.81$\pm$1.25 & \\
        LCC-$\ell_1$                & 71.24$\pm$1.31 & \Plus \\
        LCC-$\ell_2$                & 69.96$\pm$2.16 & \Plus \\
        LCC-$\ell_\infty$           & 70.92$\pm$2.04 & \Plus \\
        Dropout + Spectral Decay    & 68.97$\pm$1.31 & \\
        Dropout + LCC-$\ell_1$      & \textbf{72.18$\pm$1.97} & \Plus \\
        Dropout + LCC-$\ell_2$      & 71.15$\pm$1.13 & \Plus \\
        Dropout + LCC-$\ell_\infty$ & 70.99$\pm$1.03 & \Plus \\
        \bottomrule
    \end{tabular}
\end{table}

\begin{table}[t]
    \center
    \caption{Prediction accuracies of WRNs trained with spectral decay, dropout, batchnorm, LCC, and combinations thereof on the SINS-10 dataset. The \Plus/\Minus \, column indicates whether adding LCC to the combination of regularisers results in a statistically significant improvement or degradation in performance at the 95\% confidence level.}
    \label{tbl:sins10-wrn}
    \begin{tabular}{lcc}
        \toprule
        Method & WRN-16-4 & \Plus/\Minus \\
        \midrule
        None                        & 68.26$\pm$1.89 & \\
        Dropout                     & 68.14$\pm$2.78 & \\
        Spectral Decay              & 76.85$\pm$1.29 & \Plus \\
        LCC-$\ell_1$                & 72.85$\pm$1.63 & \Plus \\
        LCC-$\ell_2$                & 75.89$\pm$2.02 & \Plus \\
        LCC-$\ell_\infty$           & 74.09$\pm$2.19 & \Plus \\
        Dropout + Spectral Decay    & \textbf{78.57$\pm$1.37} & \Plus \\
        Dropout + LCC-$\ell_1$      & 73.27$\pm$1.19 & \Plus \\
        Dropout + LCC-$\ell_2$      & 77.93$\pm$1.19 & \Plus \\
        Dropout + LCC-$\ell_\infty$ & 76.80$\pm$1.05 & \Plus \\
        \bottomrule
    \end{tabular}
\end{table}

In contrast to the previous experiments, spectral decay is a very effective regulariser for wide residual network models trained on this dataset, with the combination of dropout and spectral decay being the best results. The hypothesis tests indicate that for both architectures, networks trained with LCC perform statistically significantly better than comparable networks trained without LCC.

\subsection{Fully Connected Networks}
Neural networks consisting exclusively of fully connected layers have a long history of being applied to classification problems arising in data mining scenarios. To evaluate how well the LCC regularisers work on tabular data, we have trained fully connected networks on the classification datasets collected by~\citet{geurts2005}. These datasets are primarily from the University California at Irvine dataset repository. The only selection criterion used by \citet{geurts2005} is that they contain only numeric features. In these experiments, each network contains two hidden layers consisting of 100 units each, and uses the ReLU activation function. Two repetitions of 5-fold cross-validation are performed for each dataset. Hyperparameters for each regulariser were tuned on a per-fold basis using grid search. The accuracy of a particular hyperparameter combination tried during the grid search was determined using a hold-out set drawn from the training data in each fold. The values considered for dropping a unit when using dropout were $p \in \{0.2, 0.3, 0.4, 0.5\}$. The values considered for $\lambda$ when using the $\ell_2$ and $\ell_\infty$ approaches were $\{2, 4, ..., 18, 20\}$, and for the $\ell_1$ variant we used $\{5, 10, ..., 45, 50\}$. Once again, the combination of LCC with each of the regularisation methods is also evaluated.

\begin{table}[t]
    \centering
	\caption{Mean test set accuracies obtained using two repetitions of 5-fold cross validation. The highest mean accuracy achieved on each dataset is bolded.}
	\label{tbl:classification}
\begin{tabular}{lcccccccc}
\toprule
 & None & DO & $\ell_1$ & $\ell_2$ & $\ell_\infty$ & DO+$\ell_1$ & DO+$\ell_2$ & DO+$\ell_\infty$ \\
\midrule
dig44     & 96.96 & 95.79 & 96.83 & 96.85 & \textbf{97.11} & 96.04                      & 96.04                      & 96.81  \\
letter    & 95.37 & 90.28 & 95.29 & 95.34 & \textbf{96.42} & 91.44 & 91.37 & 93.24 \\
pendigits & 99.44                     & 99.14 & 99.45                     & 99.45                     & \textbf{99.52}                     & 99.21 & 99.25                      & 99.41                      \\
sat       & 90.82                     & 89.18 & 90.75                     & 90.73                     & \textbf{91.00} & 90.08                      & 89.84                      & 90.06                      \\
segment   & 95.37                     & 93.70 & 95.91                     & 95.89                     & \textbf{96.52}                     & 93.90 & 93.85 & 95.52                      \\
spambase  & 94.11                     & 93.88 & 94.06                     & 93.86                     & \textbf{94.37}                                               & 93.68 & 93.68                      & 94.06                      \\
twonorm   & 97.16                     & 97.64  & 97.10                     & 97.05                     & 97.41                                               & \textbf{97.71}  & 97.68  & 97.69  \\
vehicle   & 78.02                     & 72.52 & 77.84                     & 78.07                     & \textbf{80.14}                     & 74.94                      & 74.65                      & 77.60                      \\
vowel     & 86.21 & 68.18 & 82.98                     & 83.13                     & \textbf{90.86} & 70.61                      & 71.21                      & 77.07                      \\
waveform  & 85.40                     & 86.16                      & 86.00                     & 85.82                     & 86.51                           & \textbf{86.71}  & 86.54  & 86.59 \\
\bottomrule
\end{tabular}
\end{table}

Several interesting trends can be found in this table. One particularly surprising trend is that the presence of dropout is a very good indicator of a degradation in accuracy. Interestingly, the only exceptions to this are the two synthetic datasets, where dropout is associated with an improvement in accuracy. LCC is one of the more reliable approaches to regularisation. In particular, the LCC-$\ell_\infty$ method achieves the highest mean accuracy on eight of the 10 datasets. On the other two datasets there is no substantial difference in performance between all methods. This provides strong evidence that LCC-$\ell_\infty$ is a good choice for regularisation of neural network models trained on tabular data.

\begin{figure}[t]
\center
\resizebox{\textwidth}{!}{\includegraphics[scale=1]{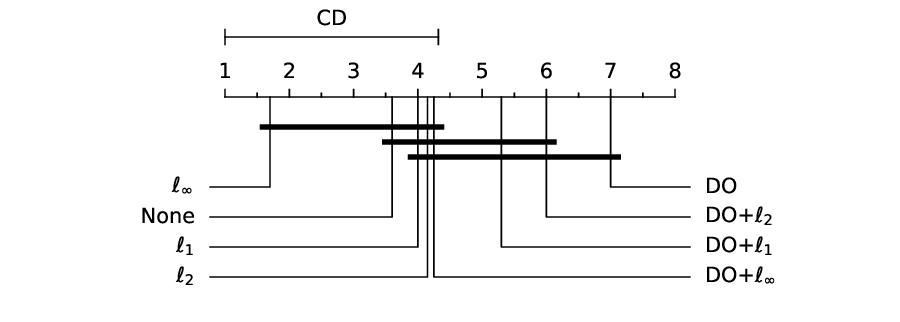}}
\caption{A critical difference diagram showing the statistically significant (95\% confidence) differences between the average rank of each method. The number beside each method is the average rank of that method across all datasets. The thick black bars overlaid on groups of thin black lines indicate a clique of methods that have not been found to be statistically significantly different.}
\label{fig:critical-diff}
\end{figure}

These results can also be visualised using a critical difference diagram~\citep{demsar2006}, as shown in Figure~\ref{fig:critical-diff}. The average rank of LCC-$\ell_\infty$ is just over 1.5, whereas the next best method---using no regularisation at all---achieves an average rank of just over 3.5. However, there is insufficient evidence to be able to state that LCC-$\ell_\infty$ statistically significantly outperforms standard neural networks. Nevertheless, it can also be seen from this diagram that LCC-$\ell_\infty$ is statistically significantly better than most of the combinations of regularisers that include dropout.

\subsection{Sensitivity to $\lambda$}
The ability to easily tune the hyperparameters of a regularisation method is important. The previous experiments have primarily taken advantage of automated hyperparameter tuning through the use of the hyperopt package~\citep{bergstra2015}, but investigating how sensitive the algorithm is to the choice of $\lambda$ could lead to useful intuition for both manual hyperparameter tuning and automated methods. The networks that have been trained with LCC regularisation thus far have required up to three different $\lambda$ hyperparameters---one for each parameterised layer type. Therefore, one cannot simply plot the model accuracy for given values of $\lambda$: it is not a scalar quantity. However, one can multiply all three of these hyperparameters by a single scalar value, and vary this scalar quantity to investigate the relationship between hyperparameter magnitude and generalisation performance. Figure~\ref{fig:sensitivity} visualises this relationship using the CIFAR-100 dataset and several models with the VGG19-style architecture. This plot was generated by defining a hyperparameter vector, $\vec \lambda = \lbrack \lambda_{conv}, \lambda_{fc}, \lambda_{bn} \rbrack$, where each component is set to the value found during the hyperparameter optimisation procedure performed as part of the experiments carried out in Section~\ref{sec:cifar100-exps}. Each data point in the plot is created by training a network with hyperparameters specified by $c \vec \lambda$, where $c$ is a user-provided scalar value, and plotting the resulting test set accuracy for different values of $c$.

\begin{figure}[t]
    \centering
    \begin{tikzpicture}
        \begin{axis}[xlabel=$c$, ylabel=Accuracy, legend pos=south east, legend style={draw=none}]
            \addplot[mark=none, color=orange] coordinates { 
                (0.8, 69.87)
                (0.85, 70.17)
                (0.9, 69.8)
                (1.0, 69.59)
                (1.2, 68.72)
                (1.4, 68.33)
                (1.6, 68.03)
                (1.8, 67.51)
            };
            \addlegendentry{LCC-$\ell_1$}
            \addplot[mark=none, color=teal] coordinates { 
                (0.8, 46.94)
                (0.85, 68.72)
                (0.9, 69.57)
                (1.0, 68.25)
                (1.2, 67.01)
                (1.4, 66.51)
                (1.6, 66.21)
                (1.8, 66.03)
            };
            \addlegendentry{LCC-$\ell_2$}
            \addplot[mark=none, color=violet] coordinates { 
                (0.8, 53.99)
                (0.85, 66.24)
                (0.9, 68.83)
                (1.0, 69.16)
                (1.2, 68.66)
                (1.4, 68.13)
                (1.6, 67.97)
                (1.8, 67.83)
            };
            \addlegendentry{LCC-$\ell_{\infty}$}
            \addplot[mark=none, dashed][domain=0.8:1.8] {65.46};
            \addlegendentry{None}
        \end{axis}
    \end{tikzpicture}
    \caption{This figure demonstrates the sensitivity of the algorithm to the choice of $\lambda$ for each of the three $p$-norms when used to regularise VGG19 networks trained on the CIFAR-100 dataset. Because a different hyperparameter was optimised for each layer type, the horizontal axis represents the value of a single constant that is used to scale the three different $\lambda$ hyperparameters associated with each curve. Note that when $c=0.6$, the LCC-$\ell_1$ network fails to converge.}
    \label{fig:sensitivity}
\end{figure}

One trend that is particularly salient in Figure~\ref{fig:sensitivity} is that choosing values for the hyperparameters that are even slightly too small results in a massive degradation in performance. Conversely, when $c$ is set above the optimal value, each method exhibits a slow decline in performance until the accuracy is comparable to that of a network trained without any regularisation. Although this is the type of behaviour one might expect from a sensible means for controlling model capacity, this second phenomenon can cause difficulty during hyperparameter tuning. It is easy to determine when the hyperparameters have been assigned values that are too small, as the model fails to converge. However, it is not easy to determine how much the hyperparameters should be increased by. It was found that for each dataset, network architecture, and $p$-norm choice, vastly different hyperparameter settings were chosen by the automated tuning process. This means there is no typical range one should expect the optimal hyperparameters to lie in, and one must use a very uninformative prior when performing hyperparameter optimisation.

\subsection{Sample Efficiency}
Imposing on a learning algorithm additional inductive biases that accurately reflect the underlying relationship between the input and output variables should result in a method that can produce well-performing models with fewer training examples than an algorithm without such inductive biases: more informative inductive biases should yield better sample efficiency. To determine if LCC improves the sample efficiency of training neural networks on image data, a series of networks are trained on progressively larger subsets of the CIFAR-10 training set. The full test set is still used for computing estimates of the accuracy of the resulting models. The learning curves for the VGG-style networks are given in Figure~\ref{fig:vgg-subsample}.

\begin{figure}[t]
    \centering
    \begin{tikzpicture}
        \begin{axis}[legend style={font=\tiny, draw=none}, legend pos=south east, ylabel=Accuracy, xlabel=Training Instances]
            \addplot[mark=none, color=black, dashed] coordinates {
                (5000, 70.14)
                (10000, 78.57)
                (15000, 82.46)
                (20000, 85.24)
                (25000, 86.5)
            };
            \addlegendentry{None}

            \addplot[mark=none, color=red, dashed] coordinates {
                (5000, 70.09)
                (10000, 78.62)
                (15000, 82.39)
                (20000, 84.88)
                (25000, 86.62)
            };
            \addlegendentry{Spectral Decay}

            \addplot[mark=none, color=blue, dashed] coordinates {
                (5000, 71.11)
                (10000, 79.28)
                (15000, 82.73)
                (20000, 84.66)
                (25000, 86.11)
            };
            \addlegendentry{Dropout}

            \addplot[mark=none, color=orange] coordinates {
                (5000, 75.1)
                (10000, 82.35)
                (15000, 84.91)
                (20000, 87)
                (25000, 88.38)
            };
            \addlegendentry{LCC-$\ell_1$}

            \addplot[mark=none, color=teal] coordinates {
                (5000, 72.98)
                (10000, 80.44)
                (15000, 84.54)
                (20000, 86.59)
                (25000, 88.08)
            };
            \addlegendentry{LCC-$\ell_2$}

            \addplot[mark=none, color=violet] coordinates {
                (5000, 74.32)
                (10000, 80.72)
                (15000, 84.45)
                (20000, 86.53)
                (25000, 87.58)
            };
            \addlegendentry{LCC-$\ell_\infty$}
        \end{axis}
    \end{tikzpicture}
    \caption{Learning curves for VGG-style networks trained on CIFAR-10 with each of the regularisation methods.}
    \label{fig:vgg-subsample}
\end{figure}

\begin{figure}
    \centering
    \begin{tikzpicture}
        \begin{axis}[legend style={font=\tiny, draw=none}, legend pos=south east, ylabel=Accuracy, xlabel=Training Instances]
            \addplot[mark=none, color=black, dashed] coordinates {
                (5000, 79.47)
                (10000, 86.36)
                (15000, 89.18)
                (20000, 91.18)
                (25000, 92.58)
            };
            \addlegendentry{None}

            \addplot[mark=none, color=red, dashed] coordinates {
                (5000, 82.02)
                (10000, 88.12)
                (15000, 90.54)
                (20000, 91.42)
                (25000, 92.53)
            };
            \addlegendentry{Spectral Decay}

            \addplot[mark=none, color=blue, dashed] coordinates {
                (5000, 80.39)
                (10000, 85.75)
                (15000, 89.49)
                (20000, 91.22)
                (25000, 92.73)
            };
            \addlegendentry{Dropout}

            \addplot[mark=none, color=orange] coordinates {
                (5000, 80.2)
                (10000, 87.64)
                (15000, 90.34)
                (20000, 92.17)
                (25000, 93.17)
            };
            \addlegendentry{LCC-$\ell_1$}

            \addplot[mark=none, color=teal] coordinates {
                (5000, 80.95)
                (10000, 87.33)
                (15000, 90.34)
                (20000, 91.96)
                (25000, 92.91)
            };
            \addlegendentry{LCC-$\ell_2$}

            \addplot[mark=none, color=violet] coordinates {
                (5000, 81.01)
                (10000, 87.08)
                (15000, 90.71)
                (20000, 92.11)
                (25000, 93.03)
            };
            \addlegendentry{LCC-$\ell_\infty$}
        \end{axis}
    \end{tikzpicture}
    \caption{Learning curves for wide residual networks trained on CIFAR-10 with each of the regularisation methods.}
    \label{fig:wrn-subsample}
\end{figure}

In the VGG plot, there is a difference of approximately 10 percentage points between the performance of the networks trained with LCC and those trained with one of the weaker baselines, for the case where only 5,000 instances are used during training. As the number of available training instances is increased, the gap between the performance of all methods becomes smaller because each method must rely less on the prior knowledge built into the learning algorithm and more on the evidence provided by the examples in the training set. Interestingly, the wide residual networks trained with the spectral decay method achieve very good performance when only a small amount of training data is available. This agrees with the previous results on the SINS-10 dataset. However, it is interesting to note that as the number of available training examples grows, this advantage is lost and the performance of networks regularised with the spectral decay method tend towards the performance of the unregularised baseline---a trend that is also noticeable in the experiments on other datasets.

\section{Conclusion}
\label{sec:conclusion}
This paper has presented a simple and effective regularisation technique for deep feed-forward neural networks called Lipschitz Constant Constraint (LCC), shown that it is applicable to a variety of feed-forward neural network architectures, and established that it is particularly suited to situations where only a small amount of training data is available. The investigation into the differences between the three $p$-norms ($p \in \{1, 2, \infty\}$) considered has provided some useful information about which one might be best-suited to the problem at hand. In particular, the $\ell_\infty$ norm appears particularly suitable for tabular data, and the $\ell_2$ norm showed the most consistently competitive performance when used as a regulariser on natural image datasets. However, given that LCC-$\ell_2$ with few power method iterations is only approximately constraining the norm, if one wants a guarantee that the Lipschitz constant of the trained network is bounded below some user-specified value, then using the $\ell_1$ or $\ell_\infty$ norm would be more appropriate.

Lastly, recent and concurrent work suggests that the utility of constraining the Lipschitz constant of neural networks is not limited to improving classification accuracy. There is already evidence that constraining the Lipschitz constant of the discriminator networks in GANs is useful~\citep{arjovsky2017,miyato2018}. Given the drawbacks in previous approaches to constraining Lipschitz constants we have outlined (cf. Section~\ref{sec:convolutional}), one might expect improvements training GANs that are $k$-Lipschitz with respect to the $\ell_1$ or $\ell_\infty$ norms, and approximately 1-Lipschitz with respect to the $\ell_2$ norm, by applying the methods presented in this paper. Exploring how well the technique presented in this paper works with recurrent neural networks would also be of interest. Finally, the experiments carried out in this paper forced all layers of the same type to have the same Lipschitz constant. This is likely an inappropriate assumption in practice, and a more sophisticated hyperparameter tuning mechanism that allows for selecting a different value of $\lambda$ for each layer could provide a further improvement to performance. However, devising a means for efficiently allocating modelling capacity on a per-layer basis is an open problem.

\bibliographystyle{plainnat}
\bibliography{paper}

\begin{thebibliography}{35}
\providecommand{\natexlab}[1]{#1}
\providecommand{\url}[1]{\texttt{#1}}
\expandafter\ifx\csname urlstyle\endcsname\relax
  \providecommand{\doi}[1]{doi: #1}\else
  \providecommand{\doi}{doi: \begingroup \urlstyle{rm}\Url}\fi

\bibitem[Arjovsky et~al.(2017)Arjovsky, Chintala, and Bottou]{arjovsky2017}
Martin Arjovsky, Soumith Chintala, and L\'eon Bottou.
\newblock Wasserstein {{GAN}}.
\newblock \emph{arXiv preprint arXiv:1701.07875}, 2017.

\bibitem[Balan et~al.(2017)Balan, Singh, and Zou]{balan2017}
Radu Balan, Maneesh Singh, and Dongmian Zou.
\newblock Lipschitz {{Properties}} for {{Deep Convolutional Networks}}.
\newblock \emph{arXiv:1701.05217 [cs, math]}, January 2017.

\bibitem[Bartlett et~al.(2017)Bartlett, Foster, and Telgarsky]{bartlett2017}
Peter~L Bartlett, Dylan~J Foster, and Matus~J Telgarsky.
\newblock Spectrally-normalized margin bounds for neural networks.
\newblock In \emph{Advances in {{Neural Information Processing Systems}} 30},
  pages 6240--6249, 2017.

\bibitem[Bartlett(1998)]{bartlett1998}
P.L. Bartlett.
\newblock The sample complexity of pattern classification with neural networks:
  The size of the weights is more important than the size of the network.
\newblock \emph{IEEE Transactions on Information Theory}, 44\penalty0
  (2):\penalty0 525--536, March 1998.

\bibitem[Bergstra et~al.(2015)Bergstra, Komer, Eliasmith, Yamins, and
  Cox]{bergstra2015}
James Bergstra, Brent Komer, Chris Eliasmith, Dan Yamins, and David~D. Cox.
\newblock Hyperopt: A {{Python}} library for model selection and hyperparameter
  optimization.
\newblock \emph{Computational Science \& Discovery}, 8\penalty0 (1):\penalty0
  014008, July 2015.

\bibitem[Cisse et~al.(2017)Cisse, Bojanowski, Grave, Dauphin, and
  Usunier]{cisse2017}
Moustapha Cisse, Piotr Bojanowski, Edouard Grave, Yann Dauphin, and Nicolas
  Usunier.
\newblock Parseval {{Networks}}: {{Improving Robustness}} to {{Adversarial
  Examples}}.
\newblock In \emph{International {{Conference}} on {{Machine Learning}}}, pages
  854--863, July 2017.

\bibitem[Dem{\v s}ar(2006)]{demsar2006}
Janez Dem{\v s}ar.
\newblock Statistical {{Comparisons}} of {{Classifiers}} over {{Multiple Data
  Sets}}.
\newblock \emph{Journal of Machine Learning Research}, 7\penalty0
  (Jan):\penalty0 1--30, 2006.

\bibitem[Gal et~al.(2017)Gal, Hron, and Kendall]{gal2017}
Yarin Gal, Jiri Hron, and Alex Kendall.
\newblock Concrete dropout.
\newblock In \emph{Advances in {{Neural Information Processing Systems}}},
  pages 3584--3593, 2017.

\bibitem[Gao and Pavel(2017)]{gao2017}
Bolin Gao and Lacra Pavel.
\newblock On the {{Properties}} of the {{Softmax Function}} with
  {{Application}} in {{Game Theory}} and {{Reinforcement Learning}}.
\newblock \emph{arXiv:1704.00805 [cs, math]}, April 2017.

\bibitem[Geurts and Wehenkel(2005)]{geurts2005}
Pierre Geurts and Louis Wehenkel.
\newblock Closed-form {{Dual Perturb}} and {{Combine}} for {{Tree}}-based
  {{Models}}.
\newblock In \emph{Proceedings of the {{22Nd International Conference}} on
  {{Machine Learning}}}, {{ICML}} '05, pages 233--240, New York, NY, USA, 2005.
  {ACM}.

\bibitem[Glorot and Bengio(2010)]{glorot2010}
Xavier Glorot and Yoshua Bengio.
\newblock Understanding the difficulty of training deep feedforward neural
  networks.
\newblock In \emph{Proceedings of the {{Thirteenth International Conference}}
  on {{Artificial Intelligence}} and {{Statistics}}}, pages 249--256, March
  2010.

\bibitem[Golowich et~al.(2018)Golowich, Rakhlin, and Shamir]{golowich2018}
Noah Golowich, Alexander Rakhlin, and Ohad Shamir.
\newblock Size-{{Independent Sample Complexity}} of {{Neural Networks}}.
\newblock In \emph{Conference {{On Learning Theory}}}, pages 297--299, July
  2018.

\bibitem[Gouk et~al.(2019)Gouk, Pfahringer, Frank, and Cree]{gouk2019}
Henry Gouk, Bernhard Pfahringer, Eibe Frank, and Michael~J. Cree.
\newblock {{MaxGain}}: {{Regularisation}} of {{Neural Networks}} by
  {{Constraining Activation Magnitudes}}.
\newblock In \emph{Machine {{Learning}} and {{Knowledge Discovery}} in
  {{Databases}}. {{ECML PKDD}} 2018.}, Lecture {{Notes}} in {{Computer
  Science}}, pages 541--556, 2019.

\bibitem[Hardt et~al.(2016)Hardt, Recht, and Singer]{hardt2016}
Moritz Hardt, Ben Recht, and Yoram Singer.
\newblock Train faster, generalize better: {{Stability}} of stochastic gradient
  descent.
\newblock In \emph{International {{Conference}} on {{Machine Learning}}}, pages
  1225--1234, June 2016.

\bibitem[He et~al.(2016)He, Zhang, Ren, and Sun]{he2016}
Kaiming He, Xiangyu Zhang, Shaoqing Ren, and Jian Sun.
\newblock Deep residual learning for image recognition.
\newblock In \emph{Proceedings of the {{IEEE}} Conference on Computer Vision
  and Pattern Recognition}, pages 770--778, 2016.

\bibitem[Ioffe and Szegedy(2015)]{ioffe2015}
Sergey Ioffe and Christian Szegedy.
\newblock Batch normalization: {{Accelerating}} deep network training by
  reducing internal covariate shift.
\newblock In \emph{International Conference on Machine Learning}, pages
  448--456, 2015.

\bibitem[Kingma and Ba(2014)]{kingma2014}
Diederik Kingma and Jimmy Ba.
\newblock Adam: {{A}} method for stochastic optimization.
\newblock \emph{arXiv preprint arXiv:1412.6980}, 2014.

\bibitem[Kingma et~al.(2015)Kingma, Salimans, and Welling]{kingma2015}
Diederik~P Kingma, Tim Salimans, and Max Welling.
\newblock Variational dropout and the local reparameterization trick.
\newblock In \emph{Advances in {{Neural Information Processing Systems}}},
  pages 2575--2583, 2015.

\bibitem[Krizhevsky and Hinton(2009)]{krizhevsky2009}
Alex Krizhevsky and Geoffrey Hinton.
\newblock \emph{Learning Multiple Layers of Features from Tiny Images}.
\newblock Master's {{Thesis}}, University of Toronto, 2009.

\bibitem[Larson(2016)]{larson2016}
Ron Larson.
\newblock \emph{Elementary Linear Algebra}.
\newblock {Nelson Education}, 2016.

\bibitem[LeCun et~al.(1998)LeCun, Bottou, Bengio, and Haffner]{lecun1998}
Yann LeCun, L\'eon Bottou, Yoshua Bengio, and Patrick Haffner.
\newblock Gradient-based learning applied to document recognition.
\newblock \emph{Proceedings of the IEEE}, 86\penalty0 (11):\penalty0
  2278--2324, 1998.

\bibitem[Miyato et~al.(2018)Miyato, Kataoka, Koyama, and Yoshida]{miyato2018}
Takeru Miyato, Toshiki Kataoka, Masanori Koyama, and Yuichi Yoshida.
\newblock Spectral {{Normalization}} for {{Generative Adversarial Networks}}.
\newblock In \emph{6th {{International Conference}} on {{Learning
  Representations}}}, 2018.

\bibitem[Neyshabur(2017)]{neyshabur2017}
Behnam Neyshabur.
\newblock \emph{Implicit {{Regularization}} in {{Deep Learning}}}.
\newblock PhD thesis, Toyota Technilogical Institute at Chicago, September
  2017.

\bibitem[Neyshabur et~al.(2018)Neyshabur, Bhojanapalli, and
  Srebro]{neyshabur2018}
Behnam Neyshabur, Srinadh Bhojanapalli, and Nathan Srebro.
\newblock A {{PAC}}-{{Bayesian Approach}} to {{Spectrally}}-{{Normalized Margin
  Bounds}} for {{Neural Networks}}.
\newblock In \emph{International {{Conference}} on {{Learning
  Representations}}}, February 2018.

\bibitem[Pugh(2002)]{pugh2002}
Charles~Chapman Pugh.
\newblock \emph{Real mathematical analysis}.
\newblock Springer, 2002.

\bibitem[Sedghi et~al.(2018)Sedghi, Gupta, and Long]{sedghi2018}
Hanie Sedghi, Vineet Gupta, and Philip~M. Long.
\newblock The {{Singular Values}} of {{Convolutional Layers}}.
\newblock \emph{arXiv:1805.10408 [cs, stat]}, May 2018.

\bibitem[Shalev-Shwartz and Ben-David(2014)]{shalev2014}
Shai Shalev-Shwartz and Shai Ben-David.
\newblock \emph{Understanding machine learning: From theory to algorithms}.
\newblock Cambridge university press, 2014.

\bibitem[Simonyan and Zisserman(2014)]{simonyan2014}
Karen Simonyan and Andrew Zisserman.
\newblock Very deep convolutional networks for large-scale image recognition.
\newblock \emph{arXiv preprint arXiv:1409.1556}, 2014.

\bibitem[Srivastava et~al.(2014)Srivastava, Hinton, Krizhevsky, Sutskever, and
  Salakhutdinov]{srivastava2014}
Nitish Srivastava, Geoffrey Hinton, Alex Krizhevsky, Ilya Sutskever, and Ruslan
  Salakhutdinov.
\newblock Dropout: {{A}} simple way to prevent neural networks from
  overfitting.
\newblock \emph{The Journal of Machine Learning Research}, 15\penalty0
  (1):\penalty0 1929--1958, 2014.

\bibitem[Tsuzuku et~al.(2018)Tsuzuku, Sato, and Sugiyama]{tsuzuku2018}
Yusuke Tsuzuku, Issei Sato, and Masashi Sugiyama.
\newblock Lipschitz-{{Margin Training}}: {{Scalable Certification}} of
  {{Perturbation Invariance}} for {{Deep Neural Networks}}.
\newblock \emph{arXiv:1802.04034 [cs, stat]}, February 2018.

\bibitem[Wan et~al.(2013)Wan, Zeiler, Zhang, Le~Cun, and Fergus]{wan2013}
Li~Wan, Matthew Zeiler, Sixin Zhang, Yann Le~Cun, and Rob Fergus.
\newblock Regularization of neural networks using dropconnect.
\newblock In \emph{International {{Conference}} on {{Machine Learning}}}, pages
  1058--1066, 2013.

\bibitem[Xiao et~al.(2017)Xiao, Rasul, and Vollgraf]{xiao2017}
Han Xiao, Kashif Rasul, and Roland Vollgraf.
\newblock Fashion-{{MNIST}}: A {{Novel Image Dataset}} for {{Benchmarking
  Machine Learning Algorithms}}.
\newblock \emph{arXiv:1708.07747 [cs, stat]}, August 2017.

\bibitem[Xu and Mannor(2012)]{xu2012}
Huan Xu and Shie Mannor.
\newblock Robustness and generalization.
\newblock \emph{Machine learning}, 86\penalty0 (3):\penalty0 391--423, 2012.

\bibitem[Yoshida and Miyato(2017)]{yoshida2017}
Yuichi Yoshida and Takeru Miyato.
\newblock Spectral {{Norm Regularization}} for {{Improving}} the
  {{Generalizability}} of {{Deep Learning}}.
\newblock \emph{arXiv:1705.10941 [cs, stat]}, May 2017.

\bibitem[Zagoruyko and Komodakis(2016)]{zagoruyko2016}
Sergey Zagoruyko and Nikos Komodakis.
\newblock Wide {{Residual Networks}}.
\newblock In \emph{Proceedings of the {{British Machine Vision Conference}}
  ({{BMVC}})}, September 2016.

\end{thebibliography}

\end{document}